\documentclass[runningheads]{llncs}
\usepackage[T1]{fontenc}
\usepackage{graphicx,verbatim}
\usepackage{booktabs}
\usepackage{multirow}
\usepackage{array,booktabs}
\usepackage[hidelinks]{hyperref}
\usepackage{amsmath}
\usepackage[table]{xcolor}

\newcolumntype{C}[1]{>{\centering\arraybackslash}p{#1}}
\usepackage{booktabs}
\usepackage{pifont}   
\usepackage{siunitx}  
\usepackage{orcidlink}
\newcommand{\cmark}{\ding{51}} 
\newcommand{\xmark}{\ding{55}} 

\usepackage[table]{xcolor}

\definecolor{bestgreen}{RGB}{102,187,106}   
\definecolor{secondgreen}{RGB}{220,237,200} 

\newcommand{\best}[1]{\bfseries #1}

\begin{document}
\title{Opportunistic Cardiac Health Assessment: Estimating Phenotypes from Localizer MRI through Multi-Modal Representations}
\titlerunning{Opportunistic Cardiac Health Assessment: Phenotypes from Localizer MRI}
%

\author{Busra Nur Zeybek\inst{1}\orcidlink{0009-0003-4755-9696} \and
Özgün Turgut\inst{1}\orcidlink{0009-0002-8704-0277} \and
Yundi Zhang\inst{1}\orcidlink{0009-0008-7725-6369} \and
Jiazhen Pan\inst{1,4}\orcidlink{0000-0002-6305-8117} \and
Robert Graf\inst{1,2}\orcidlink{0000-0001-6656-3680} \and
Sophie Starck\inst{1}\orcidlink{0000-0003-2495-6114} \and
Daniel Rueckert\inst{1,3,4}\orcidlink{0000-0002-5683-5889} \and
Sevgi Gokce Kafali\inst{1}\orcidlink{0000-0001-5941-5399}}
\authorrunning{B. N. Zeybek et al.}
\institute{
Chair for AI in Healthcare and Medicine, Technical University of Munich (TUM) \\and TUM University Hospital, Germany\\
    \and 
        Department of Diagnostic and Interventional Neuroradiology, School of Medicine, TUM University Hospital, Germany.
    \and
    Department of Computing, Imperial College London, United Kingdom\\
    \and
    Munich Center for Machine Learning (MCML), Germany\\
    \email{s.kafali@tum.de}\\
    }
\maketitle              
\vspace{-0.6cm}
\begin{abstract}
Cardiovascular diseases are the leading cause of death. Cardiac phenotypes (CPs), e.g., ejection fraction, are the gold standard for assessing cardiac health, but they’re derived from cine cardiac magnetic resonance imaging (CMR), which is costly and requires high spatio-temporal resolution. Every magnetic resonance (MR) examination begins with rapid and coarse localizers for scan planning, which are discarded thereafter. Despite non-diagnostic image quality and lack of temporal information, localizers can provide valuable structural information rapidly. In addition to imaging, patient-level information, including demographics and lifestyle, influence the cardiac health assessment. Electrocardiograms (ECGs) are inexpensive, routinely ordered in clinical practice, and capture the temporal activity of the heart. Here, we introduce \textbf{C-TRIP} (\textbf{C}ardiac \textbf{T}ri-modal \textbf{R}epresentations for \textbf{I}maging \textbf{P}henotypes), a multi-modal framework that aligns localizer MRI, ECG signals, and tabular metadata to learn a robust latent space and predict CPs using localizer images as an opportunistic alternative to CMR. By combining these three modalities, we leverage cheap spatial and temporal information from localizers, and ECG, respectively while benefiting from patient-specific context provided by tabular data. Our pipeline consists of three stages. First, encoders are trained independently to learn uni-modal representations. The second stage fuses the pre-trained encoders to unify the latent space. The final stage uses the enriched representation space for CP prediction, with inference performed exclusively on localizer MRI. Proposed C-TRIP yields accurate functional CPs, and high correlations for structural CPs. Since localizers are inherently rapid and low-cost, our C-TRIP framework could enable better accessibility for CP estimation. 
\keywords{Contrastive Learning  \and MRI \and Opportunistic Screening \and Cardiac Biomarkers.}

\end{abstract}
\section{Introduction}

Cardiovascular diseases (CVD) are the leading cause of death globally \cite{CardiovascularDiseasesCVDsb}. Several important cardiac phenotypes (CPs), such as left and right ventricular ejection fraction (LVEF, RVEF), left ventricular mass (LVM) and right ventricular end-diastolic volume (RVEDV) are established clinical biomarkers for CVD detection \cite{chenMyocardialSegmentationCardiac2022a}. Cardiac cine magnetic resonance imaging (CMR) is the reference standard for extracting structural (LVM, RVEDV) and functional CPs (LVEF, RVEF)\cite{rajiahCardiacMRIState2023c}.

However, CP calculation from CMR is a complex process\cite{rajiahCardiacMRIState2023c}. As a dynamic acquisition, CMR requires high spatio-temporal resolution and electrocardiogram (ECG) gated reconstruction to bin data from the same cardiac phases. Given the need for a dedicated full cardiac MRI protocol and additional ECG hardware for a CMR scan, there is an evident need for more efficient CP estimation. In routine clinical practice, ECG serves as a first-line cardiac assessment tool due to its low cost, rapid acquisition, and ability to capture the heart’s electrical activity \cite{siontisArtificialIntelligenceenhancedElectrocardiography2021a}. Yet, ECG cannot directly quantify CPs. A recent study proposed an ECG-based cardiac screening tool by applying contrastive learning to align the representation spaces of ECG and CMR to enrich the learned features, while predicting CPs solely on ECG during inference \cite{turgut2025unlocking}.

Beyond imaging and ECG, clinical decision making also relies on patient-level metadata, such as sex, body-mass-index (BMI), and lifestyle parameters, as cardiac structure and function are strongly influenced by these factors \cite{zhuAssociationLifestyleIncidence2022}. A previous study on multi-modal contrastive learning proved that aligning CMR and tabular data increased the CVD classification accuracy \cite{hager2023best}. This is also reflected in clinical settings, where physicians integrate information across modalities rather than interpreting each modality in isolation.

Every magnetic resonance imaging (MRI) examination starts with coarse but rapid localizer scans, which are acquired for planning purposes. Although localizers have lower image quality and are not targeted for diagnosis \cite{youReadabilityExtraspinalOrgans2021a}, they contain structural tissue information and require minimal scan and reconstruction time. Due to their simplicity and availability, they posses certain advantages over CMR (\textbf{Table \ref{tab:localizer_vs_cine}}). Localizer MRIs even by themselves represent a promising yet underexplored opportunity for opportunistic screening, as reported by \cite{bazzocchi2014localizer} for detection of osteoporotic vertebral fractures. Under a multi-modal framework, localizers have a bigger potential to be a screening tool. Thus, we introduce \textbf{C-TRIP} (\textbf{C}ardiac \textbf{T}ri-modal \textbf{R}epresentations for \textbf{I}maging \textbf{P}henotypes), a contrastive learning approach that anchors on localizer MRI and aligns its representation space with ECG-derived functional information and tabular clinical context
. In contrast to prior studies \cite{selivanovGlobalLocalContrastive,turgut2025unlocking,hager2023best,alvarez-florezDualPhaseCrossModalContrastive2026a}, our contributions are:
\begin{itemize}
    \item \textbf{Localizer-centric tri-modal alignment:} We use routinely acquired rapid and coarse localizers as the central modality for cheap spatial information, then align it with functional (ECG) and contextual (tabular) information.
    \item \textbf{Multi-modal training with single-modality inference:} Multi-modal contrastive alignment is performed during training, while CP prediction is conducted exclusively on localizer MRI at inference.
    \item \textbf{Strong performance in low-data regimes:} C-TRIP yields robust performance for both functional and structural CPs.
    \item \textbf{Opportunistic low-cost screening tool:} Rather than replacing CMR, C-TRIP acts as an opportunistic low-cost screening tool when full CMR is limited, enabling better accessibility to CP estimation. 
\end{itemize}

\begin{table}[t]
\caption{Comparison of routinely acquired localizer MRI and CMR.}
\label{tab:localizer_vs_cine}
\centering

{\fontsize{8.1}{8.1}\selectfont
\setlength{\tabcolsep}{4pt}
\renewcommand{\arraystretch}{1.15}

\begin{tabular}{@{}p{1.9cm} p{4.9cm} p{4.7cm}@{}}
\toprule
\textbf{Aspect} & \textbf{Localizer MRI} & \textbf{Cardiac CINE MRI} \\
\midrule

Purpose 
& Scan planning; not for diagnosis
& CP quantification and diagnosis \\

Scan time 
& Very short (seconds per stack) 
& Longer \\

Temporal info 
& None (static images) 
& Multi-frame across the cardiac cycle \\

Motion 
& Not ECG-gated; susceptible to motion 
& ECG-gated; captures cardiac motion \\

Image quality 
& Lower resolution; variable contrast 
& High spatial resolution and contrast \\

Standardization 
& Less standardized across sites 
& Established clinical protocols \\

Coverage 
& Coarse or incomplete coverage 
& Full short-axis + long-axis views \\

Clinical role 
& Not intended for diagnosis 
& Reference standard for CPs \\

Cost 
& Routine acquisition
& Dedicated acquisition required \\

Availability 
& Near-universal in cardiac MRI exams 
& Only during full cardiac protocol \\

\bottomrule
\end{tabular}
}
\end{table}

\begin{figure} [t]
\centering
\includegraphics[width=0.8\textwidth]{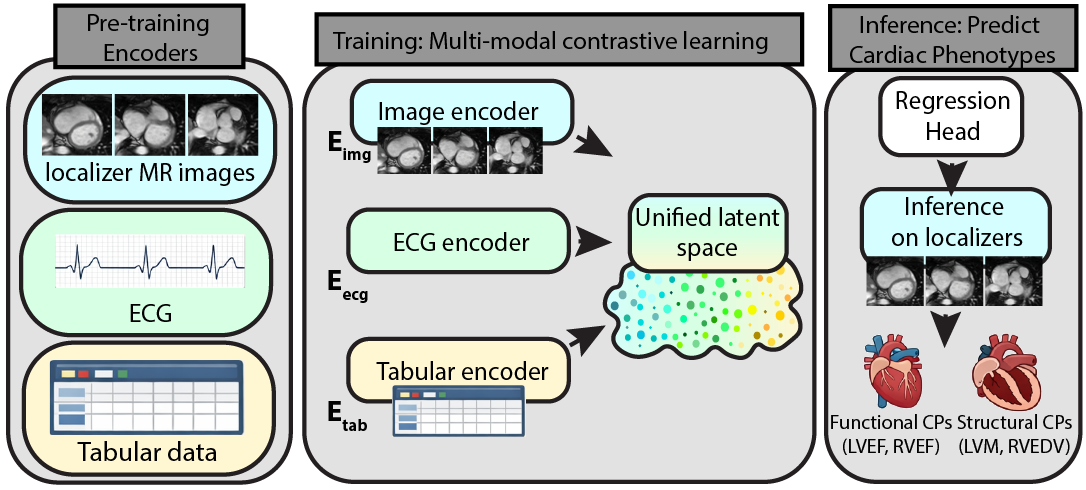}
\caption{Our multi-modal contrastive learning framework. First stage includes pre-training encoders for each modality. Second stage involves unification of the latent spaces from three modalities, in which the data from the same subjects were pulled, and different subjects were pushed away. The last stage uses a regression head to predict all 18 cardiac phenotypes solely from localizer MRIs.} \label{fig1}
\end{figure}

\section{Methods}
We propose the C-TRIP framework shown in \textbf{Fig. \ref{fig1}}. 
Our framework consists of three sequential stages. In the first stage, we pre-train three separate encoders for localizer, ECG, and tabular data using self-supervised masked data modeling. The second stage fuses the three pre-trained encoders to minimize the contrastive loss and achieve patient-level alignment in a shared latent embedding space. In the third stage, we predict CPs solely from localizers. 

\paragraph{\textbf{Stage I: Single-modality Masked Data Modeling:}} To obtain a robust latent space representation, we first train uni-modal encoders. We implement vision transformer (ViT) based masked auto encoders (MAEs)\cite{dosovitskiyImageWorth16x162021a,heMaskedAutoencodersAre2022a}, for which the architecture is adapted for each individual modality. All encoders process the 25\% visible input, while lightweight decoders reconstruct the randomly masked 75\% of the data. We minimized mean squared error (MSE) to reconstruct the masked patches. Multi-channel ECG signals are  divided into non-overlapping 1D patches similar to \cite{turgut2025unlocking}, and reconstructed to capture underlying temporal dynamics. For tabular data, the input is processed as a 1D sequence of discrete feature tokens. We project numerical and categorical features into a uniform embedding space, similar to \cite{zhang2025towards}. Full sequence is reconstructed by optimizing MSE for numerical features and cross-entropy loss for categorical features.
\paragraph{\textbf{Stage II: Multimodal Contrastive Learning:}}
We align the pre-trained uni-modal representations into a joint, unified latent space. The core objective is to enrich the anatomical features extracted from the localizers with the functional and temporal information from the ECG and the clinical context from the tabular data. To achieve this, we utilize a localizer-centric contrastive loss that aligns localizers ($L$), ECG ($E$) and tabular data ($T$), without aligning $T$ and $E$. 
\\
\textbf{Localizer-Centric Contrastive Loss:}
We independently align the localizer embeddings with the auxiliary modalities by optimizing an InfoNCE-based contrastive loss \cite{li2023scaling,radford2021learning}. Generically, let $M$ denote an auxiliary modality representing ECG ($E$) or tabular data ($T$). The loss is defined symmetrically in both directions between localizer MRI ($L$) and modality $M$. The bidirectional alignment ($\mathcal{L}_{L \leftrightarrow M}$) pulls the embeddings of the same patient closer while pushing the embeddings of different patients in the batch.

\begin{equation}
    \mathcal{L}_{L \leftrightarrow M} = \frac{1}{2} \left( \mathcal{L}_{L \rightarrow M} + \mathcal{L}_{M \rightarrow L} \right)
\end{equation}

\begin{equation}
\mathcal{L}_{L \rightarrow M} = -\frac{1}{N} \sum_{i=1}^{N} \log \frac{\exp(\text{sim}(z_L^{(i)}, z_M^{(i)}) / \tau_{LM})}{\sum_{j=1}^{N} \exp(\text{sim}(z_L^{(i)}, z_M^{(j)}) / \tau_{LM})}
\end{equation}

In the above formulation, $N$ is the batch size, $\text{sim}(\cdot, \cdot)$ denotes cosine similarity, $z_L$ and $z_M$ are the extracted representations, and $\tau_{LM}$ is a learnable temperature parameter. The overall loss objective is then 
$
\mathcal{L}_{Total} = \frac{1}{2} \left( \mathcal{L}_{L \leftrightarrow E} + \mathcal{L}_{L \leftrightarrow T} \right)
$.
By utilizing the localizer MRI as the central modality, ECG and tabular embeddings are aligned indirectly. We intentionally avoided a direct $E+T$ alignment to prevent the model from exploiting non-imaging shortcuts. This, additionally, reduced computational overhead during loss computation.
\paragraph{\textbf{Stage III: Inference to Predict CPs from Localizers:}} We attach a lightweight MLP regression head to predict CPs from localizers only during inference.

\subsection{Baselines and Ablation Studies}
To evaluate the proposed \textbf{C-TRIP} framework, we compared our approach against several baselines and conducted targeted ablation studies:\\
\textbf{Baseline networks} included supervised schemes to predict the CPs as a regression task, with different types of inputs: CMR (the reference, CMR$_{sup}$), localizers ($L_{sup}$), tabular data ($T_{sup}$), and ECG ($E_{sup}$). \newline 
\textbf{The first ablation study} analyzed a tri-modal scheme on two different conditions: tabular data injected with CPs explicitly ($L+E+T^{p}$) inspired by \cite{hager2023best}, versus tabular data without any CPs ($L+E+T$, i.e., proposed \textbf{C-TRIP}). This way, we could understand if no explicit supervision (i.e., no CP injection) could still generalize well in a wide test population.
\newline 
\textbf{The second ablation study} aimed to choose the most optimum modality combination, comparing bimodal and trimodal contrastive pre-training strategies: $L+E$, $L+T$, $L+E+T$ (i.e., our proposed \textbf{C-TRIP}).

\subsection{Dataset and Preprocessing}
We utilized paired multi-modal data from the UK Biobank \cite{raisi2021cardiovascular}, with imaging parameters detailed in \cite{petersen2016uk}. To prevent data leakage and ensure a fair evaluation, we implemented subject-level partition with demographically balanced training, validation, and test sets ($N_{train}:14577, N_{val}:3177, N_{test}:3123$). All downstream experiments strictly evaluated subjects possessing complete multi-modal pairings and 18 CPs as ground truth labels \cite{bai2020population}. For $CMR_{sup}$, we extracted 3 short-axis time frames from the middle baso-apical slice (end-diastole, end-systole, mid-phase) per subject, identical to \cite{turgut2025unlocking}. For localizer MRI, the heart was segmented using \cite{graf2025vibesegmentator} and cropped from thorax, to extract 3 consecutive oblique-transverse slices per subject. All the images were intensity normalized and resized to $224 \times 224$. Standard 10-second, 12-lead resting ECGs (500Hz) were corrected for baseline drift to eliminate respiratory and motion artifacts. Tabular data consists of numerical (e.g., BMI, metabolic rates, QRS duration) and categorical features (demographics, medical history, smoking status). Since all the modalities were recorded during the same assessment visit, they provide concurrent representation of the heart.
\subsection{Implementation}
All experiments were implemented in PyTorch/PyTorch Lightning and optimized on a single NVIDIA A40 GPU using the AdamW optimizer \cite{loshchilovDecoupledWeightDecay2019a} with cosine annealing and early stopping. In Stage I, uni-modal MAEs (embedding dimension (dim): $L:768$, $E:384$, $T:384$) were pre-trained for 400 epochs (batch size (bs) 256, learning rate (lr) $5 \times 10^{-4}$). Random augmentations like cropping, scaling, and rotation were applied only to the localizer training set.

In Stage II, representations were projected into a shared 256-dim space. Contrastive alignment was optimized for 150 epochs (bs 256) with pair-specific temperature set to $\tau_{LE} = 0.1$ and $\tau_{LT} = 0.25$ after hyperparameter tuning.

In Stage III, downstream regression head was trained for 100 epochs. To preserve the learned representations in Stage II, we applied different lr's,  $1 \times 10^{-3}$ for the regresion head, and $1 \times 10^{-4}$ for pre-trained localizer encoder.

\subsection{Evaluation} The reference CPs were extracted from CMR. Accuracy was evaluated using mean differences (MD), limits of agreement (LoA), and Pearson’s $R$ \cite{obuchowski2015quantitative}. We examined the effects of the number of fine-tuning samples during downstream analysis at stage III to mimic a realistic low-data regime (e.g., $1\%$ of the samples). 

\section{Results}

\newcolumntype{C}[1]{>{\centering\arraybackslash}p{#1}}
\newcommand{\loa}[1]{\makebox[1.50cm][c]{$#1$}}
\newcommand{\loab}[1]{\makebox[1.50cm][c]{$\pmb{#1}$}}
\newcommand{\bestloa}[1]{\cellcolor{gray!30}\makebox[1.50cm][c]{$#1$}}
\begin{table}[t]
\caption{Performance comparison across modalities. T/T$^{p}$: tabular data without/with CPs; E: electrocardiogram; L: localizer MRI. CMR$_{sup}$: provided as reference. Bold text and gray shading denote best MD and LoA.}
\label{tab:all_results}
\centering

{\fontsize{8.3}{8}\selectfont
\setlength{\tabcolsep}{1.3pt}
\renewcommand{\arraystretch}{0.98}

\begin{tabular}{@{}p{1.55cm}@{}c@{}c@{}c@{}c
S[table-format=-1.1] C{1.50cm}
S[table-format=-1.1] C{1.50cm}
S[table-format=-1.1] C{1.50cm}
S[table-format=-1.1] C{1.50cm}@{}}
\toprule
 & L & E & T$^{p}$ & T
 & \multicolumn{2}{c}{LVEF [\%]}
 & \multicolumn{2}{c}{LVM [g]}
 & \multicolumn{2}{c}{RVEF [\%]}
 & \multicolumn{2}{c}{RVEDV [mL]} \\
\cmidrule(lr){2-5}
\cmidrule(lr){6-7}
\cmidrule(lr){8-9}
\cmidrule(lr){10-11}
\cmidrule(lr){12-13}
 &  &  &  &
 & {MD} & {LoA}
 & {MD} & {LoA}
 & {MD} & {LoA}
 & {MD} & {LoA} \\
\midrule

\textit{CMR$_{\text{sup}}$} & \xmark & \xmark & \xmark & \xmark
& \bfseries 0.3  & \loab{[-8.0,8.5]}
& \bfseries -0.3 & \loab{[-24.5,24.0]}
& \bfseries -0.2 & \loab{[-8.7,8.4]}
& \bfseries 2.8  & \loab{[-39.2,44.8]}\\
\midrule

L$_{\text{sup}}$ & \cmark & \xmark & \xmark & \xmark
& \best{0.1}  & \loa{[-10.8,11.1]}
& 0.7  & \loa{[-23.4,24.8]}
& 0.1  & \loa{[-10.3,10.4]}
& 1.8  & \loa{[-39.8,43.4]}\\

E$_{\text{sup}}$ & \xmark & \cmark & \xmark & \xmark
& 0.4  & \loa{[-10.9,11.6]}
& -0.8 & \loa{[-35.9,34.3]}
& 0.2  & \loa{[-10.8,11.2]}
& -0.6 & \loa{[-60.6,59.4]}\\

T$_{\text{sup}}$ & \xmark & \xmark & \xmark & \cmark
& \best{0.1}  & \loa{[-11.3,11.5]}
& 2.3  & \loa{[-22.5,27.0]}
& -0.5 & \loa{[-11.5,10.4]}
& 5.0  & \loa{[-38.9,48.9]}\\
\midrule

L+T & \cmark & \xmark & \xmark & \cmark
& \best{0.1} & \bestloa{[-10.6,10.8]}
& 0.3 & \loa{[-23.1,23.8]}
& \best{0.0} & \bestloa{[-10.2,10.2]}
& 1.1 & \loa{[-39.8,41.9]} \\

L+E & \cmark & \cmark & \xmark & \xmark
& 0.3 & \loa{[-10.6,11.1]}
& 0.9 & \loa{[-23.5,25.2]}
& 0.2 & \loa{[-10.1,10.5]}
& 1.7 & \loa{[-39.9,43.3]} \\

L+E+T$^{p}$ & \cmark & \cmark & \cmark & \xmark
& 0.2 & \loa{[-10.5,11.0]}
& 0.9 & \bestloa{[-22.1,24.0]}
& 0.2 & \loa{[-10.0,10.5]}
& 1.2 & \bestloa{[-39.5,41.9]}  \\
\midrule

\textbf{L+E+T} & \cmark & \cmark & \xmark & \cmark
& 0.3 & \loa{[-10.4,11.1]}
& \best{0.2} & \loa{[-23.2,23.7]}
& 0.2 & \bestloa{[-10.0,10.4]}
& \best{0.2} & \loa{[-40.8,41.1]}\\
\textbf{(C-TRIP)} &  &  &  &  &  &  &  &  &  &  &  &  \\
\bottomrule

\end{tabular}
}
\end{table}

\begin{figure}[h]
\centering
\includegraphics[width=\textwidth]{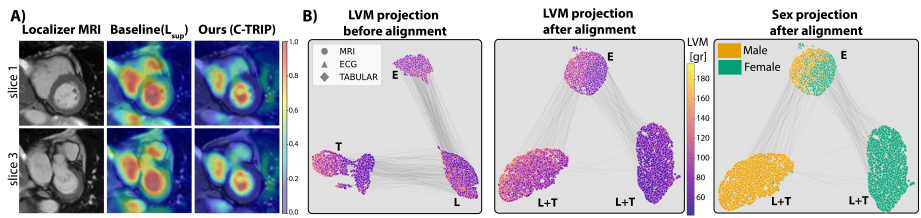}
\caption{\textbf{A) Attention maps}. Maps represent the [CLS] token's self-attention weights from the final ViT block, head-averaged and up-sampled.
While the supervised localizer baseline, $L_{sup}$, exhibits leaked attention to non-cardiac structures (e.g., lungs and chest wall), C-TRIP was able to focus on biologically relevant structures. Images are cropped to the heart region for visibility. \textbf{B) UMAPs} showing representation space before and after alignment. Our localizer-centric design reflects itself in UMAPs. Due to strong anatomical correlation, $L$ and $T$ merge easily. The relationship between $L$ and $E$ is more complex, so $E$ remains more distinct, but is pulled toward the same phenotypic gradient/trend from low to high LVM, or male/female distinction.} 
\label{fig2}
\end{figure}

\begin{figure} [h]
\centering
\includegraphics[width=.97\textwidth]{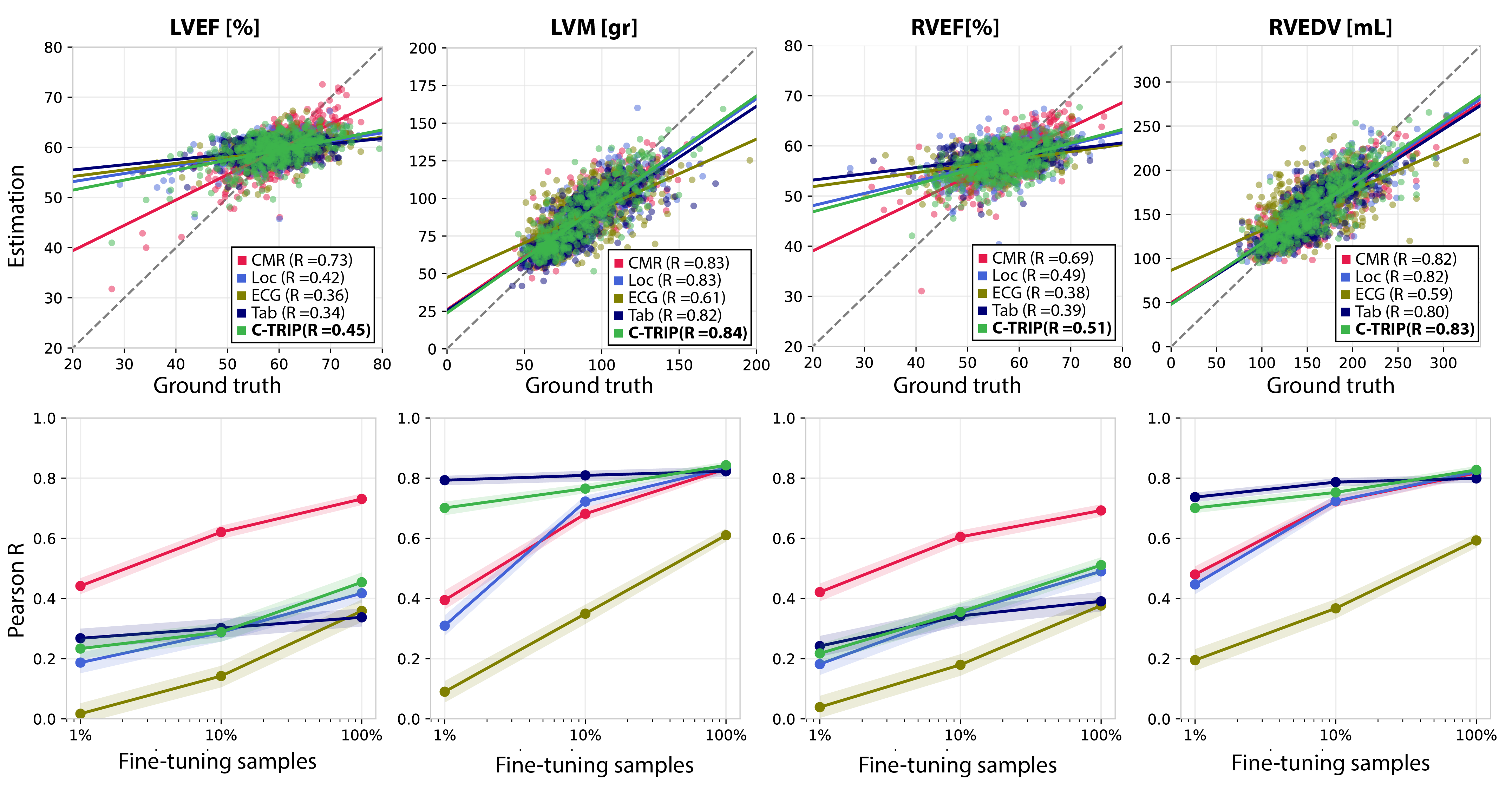}
\caption{\textbf{Performance and scaling comparison of our proposed multimodal approach against unimodal supervised baselines across four CPs.} 
    \textbf{(Top Row)} Scatter plots comparing model predictions to ground truth values using 100\% of the available fine-tuning data for LVEF, LVM, RVEF, RVEDV. Pearson $R$ is reported in the legends. The plots show 500 subjects for better visibility.
    \textbf{(Bottom Row)} Fine-tuning data scaling behavior. The plots demonstrate model performance (Pearson $R$) as a function of the fine-tuning subset size (1\%, 10\%, and 100\%, displayed on a logarithmic scale). The shaded regions represent the 95\% confidence intervals. Our proposed method C-TRIP (in green) is compared against single modality baselines.}
    \label{fig3}
\end{figure}

\textbf{Table \ref{tab:all_results}} demonstrates the CP estimation in all techniques. The CMR$_{sup}$ achieved the best performance in functional CPs, as expected, while localizer-based techniques are competitive for structural CPs. L$_{sup}$ already produces strong results (i.e., small MDs) for functional CPs. ECG$_{sup}$ yields the worst overall performance. The inclusion of explicit CPs in the tabular metadata, $L+E+T^{p}$, does not improve the performance, in fact, leads to an increase in MDs in structural CPs. Our proposed C-TRIP yields the smallest MD, with the second tightest LoA overall, showing its robust performance in structural CPs, while still being competitive in functional CPs. 

\textbf{Fig \ref{fig2}A} shows that the attention maps for C-TRIP are more focal and localized than the baseline $L_{sup}$. In slice 1 (short-axis view), C-TRIP emphasizes the \textit{donut} of the LV and the \textit{crescent} of the RV. C-TRIP shifts its attention to the specific chambers in slice 3 (a more basal/outflow view).

The multi-modal latent space (e.g., UMAPs) before and after localizer-centric alignment is shown in \textbf{Fig \ref{fig2}B} color coded by LVM and sex. $L$ and $T$ representations become deeply intertwined after alignment due to high structural  and functional correlation (e.g., males have larger LVMs, LVEF is influenced by sex). The ECG cluster remains distinct due to a more complex relationship between $L$ and $E$ since there is an inherent domain gap between electrophysiology (ECG) and anatomy (localizers). Yet, in the overall latent space, we observe a gradient from lower to higher LVM values, and male/female distinction. This particular trend indicate that the C-TRIP learned a robust mapping for CPs.

\textbf{Fig \ref{fig3}} illustrates that CMR$_{sup}$ achieves the highest correlation for functional CPs, even at low-data regimes. $L_{sup}$ show moderate performance across all data regimes. $E_{sup}$ yields the worst overall performance in both functional and structural CPs. $T_{sup}$ leads to robust performance in structural CPs, but not in functional CPs. C-TRIP achieves high correlations in structural CPs, even at $1\%$ data regime (Pearson $R>0.70$), while still performing the second-best after CMR for functional CPs. 
\section{Discussion and Conclusion}
\textbf{Proposed C-TRIP estimated CPs solely from routinely acquired localizers, to act as an opportunistic screening tool for cardiac health assessment at low-cost}. Our extensive evaluations proved the following; tabular data provides a strong population-level prior, ECG is the primary source of calibration, reducing MD. Yet, by itself, ECG performance still suffers. By aligning three modalities, C-TRIP uses the ECG to correct for generalized guess from the tabular data to consider subject-specific information, while benefiting from the spatial prior by the localizers. The impact of tri-modal alignment is most pronounced in structural CPs, particularly RV, indicating the complex geometry of the RV requires the full integration of structural, temporal and tabular priors.

\textbf{C-TRIP learned a robust anatomical prior}; only the heart-specific pixels are relevant for CP estimation, while mitigating the non-cardiac noise that confuses the baseline $L_{sup}$ as shown in attention maps. The UMAPs showed no \textit{one-blob} for C-TRIP, partly due to our localizer-centric design, partly by the the natural modality gap in CLIP \cite{liang2022mind}. While tabular data and MRI share a direct structural correlation (e.g., body scale and sex affect heart size), the relationship between ECG and MRI is more nuanced. The residual separation of the ECG cluster represents the unique electrophysiological information provided by ECG, which adds to the already learned structural information of the localizer. Our localizer-centric design also specifically prevents shortcut correlations between non-imaging features (i.e., older subjects having certain ECG rhythms). This is crucial since CPs are inherently imaging-derived biomarkers.

Pearson $R$ values reveal a division between structural and functional CPs. Structural CP estimation yielded better performance than functional CP estimation in localizer-based techniques. As expected, CMR maintains the highest correlation in functional CPs, as it captures temporal cardiac motion. In low-data regimes ($1 - 10\%$ fine-tuning), $L_{sup}$ and $E_{sup}$ suffer from low overall performance. The relatively strong performance of $T_{sup}$ in structural CPs is expected, as tabular data can extract accurate population-level priors in (e.g., males have higher LVM than females). Yet, $T_{sup}$ suffers to encode subject-specific functional/temporal information, yielding worse performance in functional CPs. C-TRIP integrates functional/temporal, contextual, and spatial information, leading to \textbf{better efficiency at low-data regimes}.

Our study had limitations. Although UK Biobank was conducted in four sites, it offers a homogeneous dataset under a single vendor, single scanner type \cite{raisi2021cardiovascular}. Additionally, conventional localizers are not acquired in a standardized manner, as opposed to UK Biobank. Thus, future work includes validating C-TRIP under less homogeneous settings. Furthermore, CMR$_{sup}$ does not consider full stack of cine, which was chosen to ensure architectural comparability across $L$ based models. Hence, it does not reflect the full diagnostic capacity of clinical CMR.

Our results support \textbf{the potential clinical utility of C-TRIP}, not as a replacement for CMR, but as a low-cost opportunistic screening tool. By enhancing rapid localizer MRI with routinely collected data (ECG and tabular), we enable reliable CP estimation when comprehensive CMR cannot be performed.

\begin{credits}
\subsubsection{\ackname} Dr. Sevgi Gokce Kafali has been sponsored by the Alexander von Humboldt Foundation. This research has been conducted using the UK Biobank Resource
under Application Number 87802.

\end{credits}

    



%
%
%
\bibliographystyle{splncs04_unsrt}
\bibliography{miccai2026}

@article{zhuAssociationLifestyleIncidence2022,
  title = {Association of {{Lifestyle With Incidence}} of {{Heart Failure According}} to {{Metabolic}} and {{Genetic Risk Status}}: {{A Population-Based Prospective Study}}},
  shorttitle = {Association of {{Lifestyle With Incidence}} of {{Heart Failure According}} to {{Metabolic}} and {{Genetic Risk Status}}},
  author = {Zhu, Zhengbao and Li, Fu-Rong and Jia, Yiming and Li, Yang and Guo, Daoxia and Chen, Jingsi and Tian, Haili and Yang, Jing and Yang, Huan-Huan and Chen, Li-Hua and Zhang, Kaixin and Yang, Pinni and Sun, Lulu and Shi, Mengyao and Zhang, Yonghong and Qin, Li-Qiang and Chen, Guo-Chong},
  year = 2022,
  month = sep,
  journal = {Circulation: Heart Failure},
  volume = {15},
  number = {9},
  issn = {1941-3289, 1941-3297},
  doi = {10.1161/CIRCHEARTFAILURE.122.009592},
  urldate = {2026-02-25},
  langid = {english}
}

@misc{CardiovascularDiseasesCVDsb,
  title = {Cardiovascular Diseases ({{CVDs}})},
  urldate = {2026-02-22},
  howpublished = {https://www.who.int/news-room/fact-sheets/detail/cardiovascular-diseases-(cvds)},
  langid = {english}
}

@article{chenMyocardialSegmentationCardiac2022a,
  title = {Myocardial {{Segmentation}} of {{Cardiac MRI Sequences With Temporal Consistency}} for {{Coronary Artery Disease Diagnosis}}},
  author = {Chen, Yutian and Xie, Wen and Zhang, Jiawei and Qiu, Hailong and Zeng, Dewen and Shi, Yiyu and Yuan, Haiyun and Zhuang, Jian and Jia, Qianjun and Zhang, Yanchun and Dong, Yuhao and Huang, Meiping and Xu, Xiaowei},
  year = 2022,
  month = feb,
  journal = {Frontiers in Cardiovascular Medicine},
  volume = {9},
  pages = {804442},
  issn = {2297-055X},
  doi = {10.3389/fcvm.2022.804442},
  urldate = {2026-02-22},
  pmcid = {PMC8914019},
  pmid = {35282363}
}

@misc{dosovitskiyImageWorth16x162021a,
  title = {An {{Image}} Is {{Worth}} 16x16 {{Words}}: {{Transformers}} for {{Image Recognition}} at {{Scale}}},
  shorttitle = {An {{Image}} Is {{Worth}} 16x16 {{Words}}},
  author = {Dosovitskiy, Alexey and Beyer, Lucas and Kolesnikov, Alexander and Weissenborn, Dirk and Zhai, Xiaohua and Unterthiner, Thomas and Dehghani, Mostafa and Minderer, Matthias and Heigold, Georg and Gelly, Sylvain and Uszkoreit, Jakob and Houlsby, Neil},
  year = 2021,
  month = jun,
  number = {arXiv:2010.11929},
  eprint = {2010.11929},
  primaryclass = {cs},
  publisher = {arXiv},
  doi = {10.48550/arXiv.2010.11929},
  urldate = {2026-02-23},
  archiveprefix = {arXiv}
}

@inproceedings{heMaskedAutoencodersAre2022a,
  title = {Masked {{Autoencoders Are Scalable Vision Learners}}},
  booktitle = {2022 {{IEEE}}/{{CVF Conference}} on {{Computer Vision}} and {{Pattern Recognition}} ({{CVPR}})},
  author = {He, Kaiming and Chen, Xinlei and Xie, Saining and Li, Yanghao and Dollar, Piotr and Girshick, Ross},
  year = 2022,
  month = jun,
  pages = {15979--15988},
  publisher = {IEEE},
  address = {New Orleans, LA, USA},
  doi = {10.1109/CVPR52688.2022.01553},
  urldate = {2026-02-23},
  copyright = {https://doi.org/10.15223/policy-029},
  isbn = {978-1-6654-6946-3},
  langid = {english}
}

@article{siontisArtificialIntelligenceenhancedElectrocardiography2021a,
  title = {Artificial Intelligence-Enhanced Electrocardiography in Cardiovascular Disease Management},
  author = {Siontis, Konstantinos C. and Noseworthy, Peter A. and Attia, Zachi I. and Friedman, Paul A.},
  year = 2021,
  journal = {Nature Reviews. Cardiology},
  volume = {18},
  number = {7},
  pages = {465--478},
  issn = {1759-5002},
  doi = {10.1038/s41569-020-00503-2},
  urldate = {2026-02-22},
  pmcid = {PMC7848866},
  pmid = {33526938}
}

@misc{loshchilovDecoupledWeightDecay2019a,
  title = {Decoupled {{Weight Decay Regularization}}},
  author = {Loshchilov, Ilya and Hutter, Frank},
  year = 2019,
  month = jan,
  number = {arXiv:1711.05101},
  eprint = {1711.05101},
  primaryclass = {cs},
  publisher = {arXiv},
  doi = {10.48550/arXiv.1711.05101},
  urldate = {2026-02-24},
  archiveprefix = {arXiv}
}

@article{youReadabilityExtraspinalOrgans2021a,
  title = {Readability of Extraspinal Organs on Scout Images of Lumbar Spine {{MRI}} According to Different Protocols},
  author = {You, Ja Yeon and Lee, Joon Woo and Seo, Jiwoon and Chai, Jee Won and Chae, Hee Dong and Kang, Heung Sik},
  year = 2021,
  month = may,
  journal = {PLoS ONE},
  volume = {16},
  number = {5},
  pages = {e0251310},
  issn = {1932-6203},
  doi = {10.1371/journal.pone.0251310},
  urldate = {2026-02-22},
  pmcid = {PMC8118512},
  pmid = {33984010}
}

@article{turgut2025unlocking,
  title={Unlocking the diagnostic potential of electrocardiograms through information transfer from cardiac magnetic resonance imaging},
  author={Turgut, {\"O}zg{\"u}n and M{\"u}ller, Philip and Hager, Paul and Shit, Suprosanna and Starck, Sophie and Menten, Martin J and Martens, Eimo and Rueckert, Daniel},
  journal={Medical Image Analysis},
  volume={101},
  pages={103451},
  year={2025},
  publisher={Elsevier}
}

@article{rajiahCardiacMRIState2023c,
  title = {Cardiac {{MRI}}: State of the Art},
  author = {Rajiah, Prabhakar Shantha and Fran{\c c}ois, Christopher J and Leiner, Tim},
  year = 2023,
  journal = {Radiology},
  volume = {307},
  number = {3},
  pages = {e223008},
  publisher = {Radiological Society of North America}
}

@article{zhang2025towards,
  title={Towards cardiac MRI foundation models: Comprehensive visual-tabular representations for whole-heart assessment and beyond},
  author={Zhang, Yundi and Hager, Paul and Liu, Che and Shit, Suprosanna and Chen, Chen and Rueckert, Daniel and Pan, Jiazhen},
  journal={Medical Image Analysis},
  pages={103756},
  year={2025},
  publisher={Elsevier}
}

@inproceedings{li2023scaling,
  title={Scaling language-image pre-training via masking},
  author={Li, Yanghao and Fan, Haoqi and Hu, Ronghang and Feichtenhofer, Christoph and He, Kaiming},
  booktitle={Proceedings of the IEEE/CVF conference on computer vision and pattern recognition},
  pages={23390--23400},
  year={2023}
}

@inproceedings{radford2021learning,
  title={Learning transferable visual models from natural language supervision},
  author={Radford, Alec and Kim, Jong Wook and Hallacy, Chris and Ramesh, Aditya and Goh, Gabriel and Agarwal, Sandhini and Sastry, Girish and Askell, Amanda and Mishkin, Pamela and Clark, Jack and others},
  booktitle={International conference on machine learning},
  pages={8748--8763},
  year={2021},
  organization={PmLR}
}

@article{petersen2016uk,
  title={UK Biobank's cardiovascular magnetic resonance protocol},
  author={Petersen, Steffen E and Matthews, Paul M and Francis, Jane M and Robson, Matthew D and Zemrak, Filip and Boubertakh, Redha and Young, Alistair A and Hudson, Sarah and Weale, Peter and Garratt, Steve and others},
  journal={Journal of cardiovascular magnetic resonance},
  volume={18},
  number={1},
  pages={8},
  year={2016},
  publisher={Elsevier}
}

@article{raisi2021cardiovascular,
  title={Cardiovascular magnetic resonance imaging in the UK Biobank: a major international health research resource},
  author={Raisi-Estabragh, Zahra and Harvey, Nicholas C and Neubauer, Stefan and Petersen, Steffen E},
  journal={European Heart Journal-Cardiovascular Imaging},
  volume={22},
  number={3},
  pages={251--258},
  year={2021},
  publisher={Oxford University Press}
}

@article{graf2025vibesegmentator,
  title={VIBESegmentator: full body MRI segmentation for the NAKO and UK Biobank},
  author={Graf, Robert and Platzek, Paul and Riedel, Evamaria Olga and Ramsch{\"u}tz, Constanze and Starck, Sophie and M{\"o}ller, Hendrik K and Atad, Matan and V{\"o}lzke, Henry and B{\"u}low, Robin and Schmidt, Carsten Oliver and others},
  journal={European Radiology},
  pages={1--15},
  year={2025},
  publisher={Springer}
}

@article{obuchowski2015quantitative,
  title={Quantitative imaging biomarkers: a review of statistical methods for computer algorithm comparisons},
  author={Obuchowski, Nancy A and Reeves, Anthony P and Huang, Erich P and Wang, Xiao-Feng and Buckler, Andrew J and Kim, Hyun J and Barnhart, Huiman X and Jackson, Edward F and Giger, Maryellen L and Pennello, Gene and others},
  journal={Statistical methods in medical research},
  volume={24},
  number={1},
  pages={68--106},
  year={2015},
  publisher={SAGE Publications Sage UK: London, England}
}

@article{liang2022mind,
  title={Mind the gap: Understanding the modality gap in multi-modal contrastive representation learning},
  author={Liang, Victor Weixin and Zhang, Yuhui and Kwon, Yongchan and Yeung, Serena and Zou, James Y},
  journal={Advances in Neural Information Processing Systems},
  volume={35},
  pages={17612--17625},
  year={2022}
}

@inproceedings{hager2023best,
  title={Best of both worlds: Multimodal contrastive learning with tabular and imaging data},
  author={Hager, Paul and Menten, Martin J and Rueckert, Daniel},
  booktitle={Proceedings of the IEEE/CVF Conference on Computer Vision and Pattern Recognition},
  pages={23924--23935},
  year={2023}
}

@article{bazzocchi2014localizer,
  title={Localizer sequences of magnetic resonance imaging accurately identify osteoporotic vertebral fractures},
  author={Bazzocchi, A and Garzillo, G and Fuzzi, F and Diano, D and Albisinni, U and Salizzoni, Eugenio and Battista, Giuseppe and Guglielmi, Giuseppe},
  journal={Bone},
  volume={61},
  pages={158--163},
  year={2014},
  publisher={Elsevier}
}

@article{bai2020population,
  title={A population-based phenome-wide association study of cardiac and aortic structure and function},
  author={Bai, Wenjia and Suzuki, Hideaki and Huang, Jian and Francis, Catherine and Wang, Shuo and Tarroni, Giacomo and Guitton, Florian and Aung, Nay and Fung, Kenneth and Petersen, Steffen E and others},
  journal={Nature medicine},
  volume={26},
  number={10},
  pages={1654--1662},
  year={2020},
  publisher={Nature Publishing Group US New York}
}

@inproceedings{selivanovGlobalLocalContrastive,
  title={Global and Local Contrastive Learning for Joint Representations from Cardiac MRI and ECG},
  author={Selivanov, Alexander and M{\"u}ller, Philip and Turgut, {\"O}zg{\"u}n and Stolt-Ans{\'o}, Nil and Rueckert, Daniel},
  booktitle={International Conference on Medical Image Computing and Computer-Assisted Intervention},
  pages={217--227},
  year={2025},
  organization={Springer}
}

@misc{alvarez-florezDualPhaseCrossModalContrastive2026a,
  title = {Dual-{{Phase Cross-Modal Contrastive Learning}} for {{CMR-Guided ECG Representations}} for {{Cardiovascular Disease Assessment}}},
  author = {{Alvarez-Florez}, Laura and {Bujalance-Gomez}, Angel and Raijmakers, Femke and {Ruiperez-Campillo}, Samuel and Kolk, Maarten Z. H. and Wiers, Jesse and Vogt, Julia and Bekkers, Erik J. and I{\v s}gum, Ivana and Tjong, Fleur V. Y.},
  year = 2026,
  month = feb,
  number = {arXiv:2602.12883},
  eprint = {2602.12883},
  primaryclass = {eess},
  publisher = {arXiv},
  doi = {10.48550/arXiv.2602.12883},
  urldate = {2026-02-25},
  archiveprefix = {arXiv}
}

\end{document}